\documentclass[lettersize,journal]{IEEEtran}

\usepackage{amsmath,amsfonts}
\usepackage{array}
\usepackage[caption=false,font=normalsize,labelfont=sf,textfont=sf]{subfig}
\usepackage{textcomp}
\usepackage{stfloats}
\usepackage{url}
\usepackage{verbatim}
\usepackage{graphicx}
\def\BibTeX{{\rm B\kern-.05em{\sc i\kern-.025em b}\kern-.08em
    T\kern-.1667em\lower.7ex\hbox{E}\kern-.125emX}}
\usepackage{balance}

\usepackage{graphics} 
\usepackage{epsfig} 
\usepackage{multicol}
\usepackage{amssymb}  
\usepackage{algorithm}
\usepackage{algorithmicx}
\usepackage{algpseudocode}
\usepackage{soul}
\usepackage{comment}
\usepackage{epstopdf}
\usepackage{hyperref}
\usepackage{multirow}
\usepackage{cite}
\usepackage{amsmath}
\usepackage{hyperref}
\usepackage{array}
\usepackage{balance}
\usepackage[flushleft]{threeparttable}

\newtheorem{lemma}{Lemma}

\newtheorem{definition}{Definition}

\begin{document}

\title{
	Velocity-Form Data-Enabled Predictive Control of Soft Robots under Unknown External Payloads}

\author{
	Huanqing Wang$^{1}$,~\IEEEmembership{Graduate Student Member,~IEEE}, 
        Kaixiang Zhang$^{1}$,
        Kyungjoon Lee$^{2}$,
        Yu Mei$^{3}$,~\IEEEmembership{Graduate Student Member,~IEEE},
        Vaibhav Srivastava$^{3}$,~\IEEEmembership{Senior Member,~IEEE},
        Jun Sheng$^{2}$,~\IEEEmembership{Member,~IEEE},
        Ziyou Song$^{4}$,~\IEEEmembership{Senior  Member,~IEEE},
	Zhaojian Li$^{1}$,~\IEEEmembership{Senior  Member,~IEEE}
	\thanks{$^{1}$Huanqing Wang, Kaixiang Zhang and Zhaojian Li are with the Department of Mechanical Engineering, Michigan State University, East Lansing, MI 48824, USA (E-mail: \{wanghu26, zhangk64, lizhaoj1\}@egr.msu.edu).}
 \thanks{$^{2}$Kyungjoon Lee and Jun Sheng are with the Department of Mechanical Engineering, University of California Riverside, CA 92521, USA. (E-mail:\
    {klee449, jun.sheng\}@ucr.edu}.}
	\thanks{$^{3}$Yu Mei and Vaibhav Srivastava are with the Department of Electrical and Computer Engineering, Michigan State University, East Lansing, MI 48824, USA (E-mail: \{meiyu1, vaibhav\}@egr.msu.edu).}
     \thanks{$^{4}$Ziyou Song is with the Department of Electrical and Computer Engineering, University of Michigan, Ann Arbor, USA. (E-mail:
    ziyou@umich.edu).}
}%

\maketitle

\begin{abstract}
	Data-driven control methods such as data-enabled predictive control (DeePC) have shown strong potential in efficient control of soft robots without explicit parametric models. However, in object manipulation tasks, unknown external payloads and disturbances can significantly alter the system dynamics and behavior, leading to offset error and degraded control performance. In this paper, we present a novel velocity-form DeePC framework that achieves robust and optimal control of soft robots under unknown payloads. The proposed framework leverages input-output data in an incremental representation to mitigate performance degradation induced by unknown payloads, eliminating the need for weighted datasets or disturbance estimators. We validate the method experimentally on a planar soft robot and demonstrate its superior performance compared to standard DeePC in scenarios involving unknown payloads.	
\end{abstract}

    

\section{Introduction}

Soft robots, built from flexible materials such as fabrics, polymers, and elastomers, have recently gained significant attention for their ability to operate in scenarios where traditional rigid robots are inadequate. Their inherent compliance allows safe human interaction and the manipulation of delicate objects~\cite{Lee2024,Liu2024,Xavier2022,Armanini2023}. Actuated by various methods, such as pneumatic, hydraulic, cable, or magnetic systems~\cite{Wang2021}, soft robots exhibit complex nonlinear dynamics that pose substantial challenges for accurate modeling and control.

Traditionally, modeling approaches for soft robots are classified into numerical and analytical methods~\cite{ScheggDuriez2022}. Numerical methods, including finite element analysis (FEA)~\cite{Duriez2013} and Cosserat rod theory~\cite{Jones2009}, can provide accurate models but are computationally intensive, making them unsuitable for real-time control. Analytical methods usually rely on the Piecewise Constant Curvature (PCC) assumption~\cite{Webster2010,Hannan2003} and offer simplified representations suitable for quasi-static control. However, PCC-based models may lack accuracy in dynamic scenarios involving external payloads. For dynamic modeling, augmented rigid-link models~\cite{Katzschmann2019} and Lagrangian formulations~\cite{Falkenhahn2014,Cheng2024} are often employed in combination with parameter identification experiments.

In recent years, data-driven approaches have been widely applied to soft robots, including reinforcement learning (RL) methods~\cite{Thuruthel2019,Satheeshbabu2019}, Koopman operator-based models~\cite{Bruder2021TRO,Wang2023,WangJiajin2025}, and \textbf{d}ata-\textbf{e}nabl\textbf{e}d \textbf{p}redictive \textbf{c}ontrol (DeePC)~\cite{Wang2025}. ~\cite{Satheeshbabu2019} presented a model-free RL approach for open-loop control of a fiber-enforced soft bending actuator, while ~\cite{Thuruthel2019} introduced a model-based RL for closed-loop control of a soft manipulator. These RL-based methods generally require large datasets and expensive hardware for training, and their stability properties are not guaranteed. Koopman operator-based methods have gained popularity due to their ability to lift nonlinear dynamics into a higher-dimensional linear representation suitable for integration with model predictive control (MPC). However, their effectiveness depends strongly on careful selection of basis functions. By contrast, DeePC has emerged as a powerful alternative. Unlike RL, it does not require extensive offline training, and unlike Koopman methods, it does not rely on basis function selection. Instead, DeePC directly exploits input–output trajectory data, providing a model-free yet control-oriented framework. It has been successfully applied across a variety of systems.
For instance, our prior work~\cite{Wang2025} applied DeePC for soft robot control in the absence of payloads, demonstrating its advantages in handling complex, nonlinear behaviors without explicit modeling. 

Handling external payloads is one of the most critical and challenging aspects of data-driven control for soft robots, as manipulation tasks frequently involve objects of varying or unknown weights. The data-driven control methods mentioned above cannot directly accommodate additional payloads and require further modifications. For instance, Cao et al.~\cite{Cao2021} utilized a neural-network-based controller combined with echo state Gaussian processes to achieve robust soft robot tracking under varying payloads. Ji et al.~\cite{Ji2025} combined the Koopman operator with reinforcement learning to handle external payloads. However, the payloads tested in both studies were relatively small (all below 70~g). Other solutions have treated loading conditions as constant disturbances and incorporated them into Koopman-lifted state representations~\cite{Bruder2021}. 
Despite these advances, most methods still heavily rely on pre-collected datasets that explicitly include loaded conditions. This dependence requires large and diverse datasets to cover different payload scenarios, and performance cannot be guaranteed for unseen payloads, which limits the practicality of these approaches in real-world scenarios.

A promising direction to address this challenge is the offset-free MPC scheme, which is designed to handle modeling uncertainties and external disturbances. Offset-free design can be achieved by incorporating a disturbance model or by adopting incremental/velocity-form models~\cite{Pannocchia2015}. For example, \cite{CHEN2022102871} combined offset-free MPC with the Koopman operator to perform task-space tracking under model uncertainties and disturbances, and validated its robust performance when part of the robot was physically constrained. However, their experiments did not account for payloads. So far, control of soft robots under payloads has not been thoroughly investigated in direct data-driven settings. This gap motivates the need for a direct data-driven control framework that can achieve offset-free tracking and robust performance under unknown payloads without requiring pre-collected loaded data.

To address this challenge, we propose a velocity-form DeePC (VDeePC) framework for soft robots. Building on the concept of velocity-form MPC, our approach integrates the robustness of the velocity form with the advantages of DeePC. This framework enables accurate tracking and improved robustness under varying payload conditions, as demonstrated through extensive experimental validation.
The main contributions of this paper are threefold. First, we introduce a novel velocity-form DeePC framework that improves control accuracy for soft robots. To the best of our knowledge, this is the first application of the velocity-form DeePC for soft robot control, addressing both unloaded and payload cases. 
Second, different from existing data-driven approaches that typically require collecting large and diverse datasets to handle payload variations, the proposed method only requires data collected under unloaded conditions and can effectively generalize to online control with varying payloads, thereby improving its practicality in real-world scenarios. 
Third, we provide extensive experimental evaluations that compare velocity-form DeePC with standard DeePC across multiple scenarios, demonstrating its superior performance in both unloaded and payload cases.

The remainder of this paper is structured as follows.  
Section~\ref{sec_sys} describes the soft robot system and provides the theoretical background on non-parametric representations.  
Section~\ref{sec_velocity_form_DeePC} presents the formulation of DeePC and introduces our velocity-form DeePC framework.  
Section~\ref{sec_experiments} details the experimental setup and evaluates the performance of the proposed approach against standard DeePC on a soft robot through various tests.  
Finally, Section~\ref{sec_conclusion} concludes the paper.

\section{System Description and Preliminaries} \label{sec_sys}
This section begins with a description of the soft robot system, followed by a review of the preliminaries on non-parametric system representations.

    \begin{figure}[!h]
		\centering
		\includegraphics[width=0.9\linewidth]{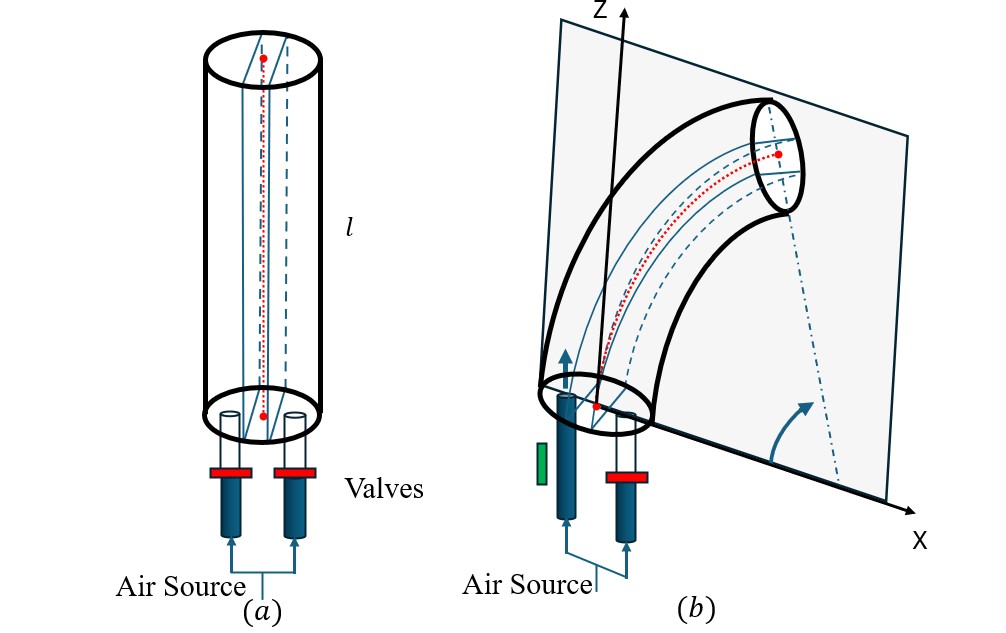}
		\caption{Illustration diagram of the soft robot, which consists of two chambers (left and right). Actuating one chamber with air produces a bending motion.  
        The geometric centerline is shown in red.}\label{fig_sys_desc}
	\end{figure}


\subsection{System Description} 
\label{sub_system_descripton}

A schematic of the soft robot is shown in Fig.~\ref{fig_sys_desc}. 
The robot uses a pneumatic soft bending actuator that consists of two chambers, one on the left and one on the right, each connected to an air tube. 
Electronic valves control the air supply to each chamber, which has a non-extensible inner layer and an extensible outer layer. 
When one chamber is pressurized, the actuator bends toward the opposite side, generating the desired bending motion. 
For details on the design and fabrication process, interested readers can refer to~\cite{Wang2025}. The geometric centerline of the actuator, highlighted in red, is often used in analytical modeling approaches such as the PCC method, where curvature is employed to compute the robot's pose parameters. In contrast, our data-driven approach directly utilizes the position of the tip (end effector) of the soft robot as the basis for control.

 \subsection{ Non-Parametric Representation} 
\label{sub_representation}

For a discrete-time linear time-invariant (LTI) system, the parametric representation can be expressed as
\begin{equation}\label{eq:parametric_description}
    \begin{aligned}
        x_{k+1} &= Ax_k + Bu_k, \\
        y_k &= Cx_k + Du_k,
    \end{aligned}
\end{equation}
where \(A \in \mathbb{R}^{n \times n}\), \(B \in \mathbb{R}^{n \times m}\), \(C \in \mathbb{R}^{p \times n}\), and \(D \in \mathbb{R}^{p \times m}\) are system matrices, while \(x_k \in \mathbb{R}^n\), \(u_k \in \mathbb{R}^m\), and \(y_k \in \mathbb{R}^p\) represent the state, control input, and output, respectively.

\begin{definition} \label{def:pe}
    For positive integers $L, T\in \mathbb{Z}$ with $L\le T$, the Hankel matrix of depth $L$ for a signal sequence $\omega_{\left[0, T-1\right]}\!:=\!\begin{bmatrix}
	  \omega^{\top}_0, \omega^{\top}_1, \cdots, \omega^{\top}_{T-1}
    \end{bmatrix}^{\top}$ is defined as
    \begin{equation}
    \mathcal{H}_{L}(\omega_{\left[0, T-1\right]}):=
	\begin{bmatrix}
		\omega_0 & \omega_1 & \cdots & \omega_{T-L} \\
		\omega_1 & \omega_2 & \cdots & \omega_{T-L+1}\\
		\vdots & \vdots & \ddots & \vdots \\
		\omega_{L-1} & \omega_{L} & \cdots & \omega_{T-1}
	\end{bmatrix}.
    \end{equation}
	The sequence $\omega_{[0,T-1]}$ is said to be \textit{persistently exciting of order $L$} 
	if the Hankel matrix $\mathcal{H}_{L}(\omega_{[0,T-1]})$ has full row rank. 
\end{definition}

The non-parametric representation is grounded in Willems' Fundamental Lemma~\cite{WILLEMS2005325}, which states that all feasible trajectories of an LTI system can be constructed from a sufficiently rich past input-output dataset, provided the input sequence is persistently exciting. This enables the system to be represented via a finite collection of its input-output data rather than explicit system matrices. 
Specifically, the input and output sequences are collected as follows:  
\begin{equation} \label{equ:ud_yd}
    \begin{aligned}
    u_{\left[0, T-1\right]}^{\mathrm{d}}&:=\begin{bmatrix}
	(u^{\mathrm{d}}_0)^{\top}, (u^{\mathrm{d}}_1)^{\top}, \cdots, (u^{\mathrm{d}}_{T-1})^{\top}
    \end{bmatrix}^{\top},
   \\
   y_{\left[0, T-1\right]}^{\mathrm{d}}&:=\begin{bmatrix}
	(y^{\mathrm{d}}_0)^{\top}, (y^{\mathrm{d}}_1)^{\top}, \cdots, (y^{\mathrm{d}}_{T-1})^{\top}
    \end{bmatrix}^{\top}.
    \end{aligned}
\end{equation}
The Hankel matrices $\mathcal{H}_L(u_{[0, T-1]}^{\mathrm{d}})$ and $\mathcal{H}_L(y_{[0, T-1]}^{\mathrm{d}})$ can be constructed from the input-output trajectories, as follows:
\begin{equation}\label{eq:non_parametric_description}
    \begin{aligned}
        \begin{bmatrix}
            \mathcal{H}_L(u^{\mathrm{d}}_{[0, T-1]}) \\
            \hline
            \mathcal{H}_L(y^{\mathrm{d}}_{[0, T-1]})
        \end{bmatrix}
        &:= 
        \begin{bmatrix}
            u^{\mathrm{d}}_0 & u^{\mathrm{d}}_1 & \dots & u^{\mathrm{d}}_{T-L} \\
            \vdots & \vdots & \ddots & \vdots \\
            u^{\mathrm{d}}_{L-1} & u^{\mathrm{d}}_L & \dots & u^{\mathrm{d}}_{T-1} \\
            \hline
            y^{\mathrm{d}}_0 & y^{\mathrm{d}}_1 & \dots & y^{\mathrm{d}}_{T - L} \\
            \vdots & \vdots & \ddots & \vdots \\
            y^{\mathrm{d}}_{L - 1} & y^{\mathrm{d}}_{L} & \dots & y^{\mathrm{d}}_{T - 1} \\
        \end{bmatrix}.
    \end{aligned}
\end{equation}

Each column in these matrices corresponds to a length $L$ input-output trajectory of the LTI system. When the input sequence is persistently exciting, the column space of the Hankel matrices serves as a non-parametric representation of the system behavior. Here, ``persistently exciting'' implies that the input signal is sufficiently rich and of adequate duration to generate outputs that fully capture the inherent behavior of the underlying system. The results are concluded in the following lemma.

\begin{lemma}[Fundamental Lemma \cite{WILLEMS2005325}] \label{lemma 1}
     Consider a controllable LTI system with an input sequence $u_{[0,T-1]}^{\mathrm{d}}$ that is persistently exciting of order $n + L$ (see Definition~\ref{def:pe}). Any length $L$ sequence $(u_{[0,L-1]}, y_{[0,L-1]})$ is a valid input-output trajectory of the system if and only if
    \begin{equation} \label{eq:fundamental-lemma}
	\begin{bmatrix}
		u_{[0,L-1]} \\ y_{[0,L-1]}
	\end{bmatrix} = \begin{bmatrix}
		\mathcal{H}_{L}(u^{\mathrm{d}}_{[0,T-1]}) \\
		\mathcal{H}_{L}(y^{\mathrm{d}}_{[0,T-1]})
	\end{bmatrix} g
	\end{equation}
    for some vector $g \in \mathbb{R}^{(T - L + 1)}$.
\end{lemma}

The above lemma establishes the relationship between the parametric and non-parametric representations, showing that all valid trajectories of the dynamical system~\eqref{eq:parametric_description} can be generated using Hankel matrices when the input is persistently exciting. With this foundation, we now proceed to formulate the DeePC problem.

\section{Control Methods} \label{sec_velocity_form_DeePC}
In this section, we begin by introducing the standard DeePC formulation, followed by its regularized variant, and then present our proposed velocity-form DeePC along with its extensions for enhanced robustness and computational efficiency.

\subsection{Data-Enabled Predictive Control} 
\label{sub_lin_deepc}
Let \(T_{\text{ini}}\in \mathbb{Z}\) denote the time horizon of ``past data'' and \(N \in \mathbb{Z}\) represent the prediction horizon for ``future data''. We define \(L = T_{\mathrm{ini}} + N\) as the total horizon. As shown in~\eqref{equ:ud_yd}, a sufficiently long input sequence \(u_{[0, T-1]}^{\mathrm{d}}\) of length \(T\), persistently exciting of order \(n + L\), is selected, and the corresponding output of the system is \(y_{[0, T-1]}^{\mathrm{d}}\). The Hankel matrices are constructed and partitioned as follows: 
\begin{equation}\label{eq:UP_Uf}
\begin{aligned}
    \begin{bmatrix}
        U_p \\ U_f
    \end{bmatrix}
    &= \mathcal{H}_L(u_{[0, T-1]}^{\mathrm{d}}), \\
    \begin{bmatrix}
        Y_p \\ Y_f
    \end{bmatrix}
    &= \mathcal{H}_L(y_{[0, T-1]}^{\mathrm{d}}),
\end{aligned}
\end{equation}
where \(U_p\) and \(U_f\) are the first \(T_{\mathrm{ini}}\) block rows and the last \(N\) block rows of \(\mathcal{H}_L(u_{[0, T-1]}^{\mathrm{d}})\), representing ``past'' control data section and ``future'' control data section, respectively. Similarly, \(Y_p\) and \(Y_f\) denote the ``past'' and ``future'' output data from \( \mathcal{H}_L(y_{[0, T-1]}^{\mathrm{d}}) \). 

For an LTI system, the DeePC optimization problem is formulated as~\cite{Coulson2019}:
\begin{equation}\label{eq:lti_deepc_formulation}
    \begin{aligned}
        \min_{g, u, y} \quad & \|y - y^r\|_Q^2  + \|u\|_R^2 \\
        \text{subject to} \quad & \begin{bmatrix} U_p \\ U_f \\ Y_p \\ Y_f \end{bmatrix} g = \begin{bmatrix} u_{\text{ini}} \\ u \\ y_{\text{ini}} \\ y \end{bmatrix}, \quad u \in \mathcal{U}, \quad y \in \mathcal{Y}.
    \end{aligned}
\end{equation}
In~\eqref{eq:lti_deepc_formulation}, \(Q\) and \(R\) are positive definite weighting matrices for the output and input, respectively.  
\(\mathcal{U}\) and \(\mathcal{Y}\) denote the input and output constraint sets. 
The past input and output sequences are \( u_{\mathrm{ini}} = u_{[t-T_{\mathrm{ini}}, t-1]} \) and \( y_{\mathrm{ini}} = y_{[t-T_{\mathrm{ini}}, t-1]} \), spanning the past horizon up to time \( t-1 \), while \( u = u_{[t, t+N-1]} \) and \( y = y_{[t, t+N-1]} \) represent the predicted input and output sequences over the horizon \( N \).
The desired trajectory is given by \(y^r = [y_{t}^{r\top}, y_{t+1}^{r\top}, \dots, y_{t+N-1}^{r\top}]^{\top}\). The optimization problem is solved in a receding-horizon fashion, and the optimal control sequence is given by \(u^* = [u^{*\top}_0, \dots, u^{*\top}_{N-1}]^{\top}\).

\subsection{Regularized DeePC for Nonlinear Systems} 
\label{sub_deepc}
The DeePC formulation in~\eqref{eq:lti_deepc_formulation} is designed for LTI systems. For stochastic or nonlinear systems, slack variables and regularization techniques can be incorporated to enhance robustness against noise and nonlinear effects. The resulting regularized DeePC formulation can be expressed as follows:
\begin{equation}\label{eq:deepc_formulation}
    \begin{aligned}
        \min_{g, u, y, \sigma_y} \quad & \|y - y^r\|_Q^2  + \|u\|_R^2  + \lambda_y \|\sigma_y\|_2^2 + \lambda_g\|g\|_2^2 \\
        \text{subject to} \quad & \begin{bmatrix} U_p \\ U_f \\ Y_p \\ Y_f \end{bmatrix} g = \begin{bmatrix} u_{\text{ini}} \\ u \\ y_{\text{ini}} \\ y \end{bmatrix} + \begin{bmatrix} 0 \\ 0 \\ \sigma_y \\ 0 \end{bmatrix}, \quad u \in \mathcal{U}, \quad y \in \mathcal{Y},
    \end{aligned}
\end{equation}
where \(\sigma_y\) is a slack variable, and \(\lambda_y \geq 0\), \(\lambda_g \geq 0\) are regularization parameters.
These terms are introduced to mitigate the effects of corruption, noise, and model mismatch~\cite{Coulson2019}. 

The steps for applying regularized DeePC to the soft robot are summarized in Algorithm~\ref{algo:deepc}. At the beginning of the process, $u_{\mathrm{ini}}$ and $y_{\mathrm{ini}}$ are initialized by setting the pressure input to zero while recording the robot's stationary position data. Once $u_{\mathrm{ini}}$ and $y_{\mathrm{ini}}$ are populated, the optimization problem is solved to obtain the optimal control sequence $u^{*}$, and only the first control input of the optimal sequence is applied.  The horizon is then shifted forward, and the process repeats.
\begin{algorithm} 
  \caption{Regularized DeePC Algorithm for Soft Robot} \label{algo:deepc}
  \begin{algorithmic}[1] %
    \State \textbf{Input:} Pre-collected pressure input $u^{\mathrm{d}}_{[0, T-1]}$, soft robot position output $y^{\mathrm{d}}_{[0, T-1]}$, final step $t_{\mathrm{end}}$.
    \State Construct Hankel matrices and partition $U_p$, $U_f$, $Y_p$, $Y_f$.
    \State For $t<T_{\text{ini}}$, initialize $u_{\text{ini}}$ with 0 and $y_{\text{ini}}$ with real-time position measurements.
    \While {$T_{\text{ini}}\le t \le t_{\mathrm{end}}$} \label{algo:line:startwhile}
      \State Solve the optimization problem~\eqref{eq:deepc_formulation} to obtain optimal pressure control sequence $u^{*}$.
      \State  Send the first step optimal pressure control $u^{*}_0$ to low-level pneumatic controller.
      \State  Measure the soft robot position, and update $u_{\mathrm{ini}}$ and $y_{\mathrm{ini}}$ to the $T_{\mathrm{ini}}$ most recent input/output measurements.
      \State  Set $t$ to $t+1$.
    \EndWhile \label{algo:line:endwhile}
  \end{algorithmic}
\end{algorithm}

\subsection{Velocity-Form DeePC} 
\label{sub_velocity_form_deepc}
For an LTI system subject to state disturbance \(w_{x,k}\) and output disturbance \(w_{y,k}\), the dynamics can be formulated as follows:
\begin{equation}\label{eq:linear_syst_dist}
\begin{aligned}
x_{k+1} &= Ax_k + Bu_k + w_{x,k}, \\
y_k &= Cx_k+w_{y,k}.
\end{aligned}
\end{equation}
The velocity-form model~\cite{Pannocchia2001,Wang2004} can be employed to address the disturbance.  
This model is obtained from the nominal system by defining the state, control, and output increments:
\begin{equation}
\begin{aligned}
    \Delta x_k &= x_{k+1} - x_k, \\
    \Delta u_k &= u_{k+1} - u_k, \\
    \Delta y_k &= y_{k+1} - y_k.
\end{aligned}
\end{equation}
Following the above definitions, the system in~\eqref{eq:linear_syst_dist} can be reformulated as follows:
\begin{equation}
\begin{aligned}
\Delta x_{k+1} &= A \Delta x_k + B \Delta u_k + \Delta w_{x,k}, \\
\Delta y_{k+1} &= C \Delta x_{k+1} + \Delta w_{y,k}, \\
               &= CA \Delta x_k + CB \Delta u_k + C \Delta w_{x,k} + \Delta w_{y,k},
\end{aligned}
\end{equation}
where $\Delta w_{x,k} = w_{x,k+1}-w_{x,k}$ and $\Delta w_{y,k} = w_{y,k+1}-w_{y,k}$.
If disturbances are constants (i.e., $\Delta w_{x,k} = 0$, $\Delta w_{y,k} = 0$), they vanish in velocity form, eliminating the need for explicit disturbance modeling and allowing the state-space model to be directly applied within the MPC framework~\cite{Pannocchia2015}. This inherent offset-free property motivates our velocity-form DeePC approach. 

In this work, we present a velocity-form DeePC grounded in Willems' Fundamental Lemma. The proposed approach embeds the velocity-form representation directly into the DeePC constraints while preserving the original cost function structure of standard DeePC. This formulation achieves robust performance in the presence of disturbances, while remaining consistent with the established DeePC framework.

We now introduce the velocity-form DeePC. By assuming constant disturbances, the system becomes:
\begin{equation}\label{eq:offset}
\begin{aligned}
    \Delta x_{k+1} &= A \Delta x_k + B \Delta u_k, \\
    \Delta y_k &= C \Delta x_k.
\end{aligned}
\end{equation}
Given pre-collected input and output sequences $u_{\left[0, T-1\right]}^{\mathrm{d}}$ and $y_{\left[0, T-1\right]}^{\mathrm{d}}$, their corresponding incremental trajectories of length $T-1$ are defined as
\begin{equation} \label{equ:Delta}
\begin{aligned}
\Delta u_{\left[0, T-2\right]}^{\mathrm{d}} &:= u_{\left[1, T-1\right]}^{\mathrm{d}} - u_{\left[0, T-2\right]}^{\mathrm{d}},
\\
\Delta y_{\left[0, T-2\right]}^{\mathrm{d}} &:= y_{\left[1, T-1\right]}^{\mathrm{d}} - y_{\left[0, T-2\right]}^{\mathrm{d}}.
\end{aligned}
\end{equation}
The associated Hankel matrices for $\Delta u_{\left[0, T-2\right]}^{\mathrm{d}}$ and $\Delta y_{\left[0, T-2\right]}^{\mathrm{d}}$ are constructed as
\begin{equation}\label{eq:delta_UP_Uf}
\begin{aligned}
    \begin{bmatrix}
        \Delta U_p \\ \Delta U_f
    \end{bmatrix}
    &= \mathcal{H}_{L-1}(\Delta u_{[0, T-2]}^{\mathrm{d}}), \\
    \begin{bmatrix}
        \Delta Y_p \\ \Delta Y_f
    \end{bmatrix}
    &= \mathcal{H}_{L-1}(\Delta y_{[0, T-2]}^{\mathrm{d}}),
\end{aligned}
\end{equation}
where \(\Delta U_p\) and \(\Delta U_f\) denote the first \(T_{\mathrm{ini}}-1\) and last \(N\) block rows of \(\mathcal{H}_{L-1}(\Delta u_{[0, T-2]}^{\mathrm{d}})\), respectively. Similarly, \( \Delta Y_p \) and \( \Delta Y_f \) correspond to the first \( T_{\mathrm{ini}}-1 \) and last \( N \) block rows of \( \mathcal{H}_{L-1}(\Delta y_{[0, T-2]}^{\mathrm{d}}) \). The past incremental input and output sequences are denoted by \( \Delta u_{\mathrm{ini}} = \Delta u_{[t-T_{\mathrm{ini}}, t-2]} \) and \( y_{\mathrm{ini}} = y_{[t-T_{\mathrm{ini}}, t-2]} \), each of length $T_{\mathrm{ini}}-1$. The predicted input and output sequences over the horizon \( N \) are \( \Delta u = \Delta u_{[t-1, t+N-2]} \) and \( \Delta y = \Delta y_{[t-1, t+N-2]} \). According to Willems' Fundamental Lemma~\cite{WILLEMS2005325}, it can be concluded that if $\Delta u_{\left[0, T-2\right]}^{\mathrm{d}}$ is persistently exciting of order $n+L-1$, then $(\Delta u_{\mathrm{ini}}, \Delta u, \Delta y_{\mathrm{ini}}, \Delta y)$ is a feasible input-output trajectory of system~\eqref{eq:offset} if and only if there exists some vector $g\in \mathbb{R}^{T-L+1}$ such that 
\begin{equation} \label{equ:Delta_U_Y}
\begin{bmatrix}
\Delta U_p \\
\Delta U_f \\
\Delta Y_p \\
\Delta Y_f
\end{bmatrix}
g =
\begin{bmatrix}
\Delta u_{\mathrm{ini}} \\
\Delta u \\
\Delta y_{\mathrm{ini}} \\
\Delta y
\end{bmatrix}.
\end{equation}

To facilitate the formulation of the velocity-form DeePC, the following variables are introduced:
\begin{equation} \label{equ:Delta_tilde_U_Y}
\begin{aligned}
\Delta \tilde{U}_p 
&= 
\underbrace{
\begin{bmatrix}
I_{m \times m} & 0 & 0 & \cdots & 0 \\
I_{m \times m} & I_{m \times m} & 0 & \cdots & 0 \\
\vdots & \vdots & \ddots & \ddots & \vdots \\
I_{m \times m} & I_{m \times m} & I_{m \times m} & \cdots & I_{m \times m}
\end{bmatrix}}
_{Q_{U_p}\in \mathbb{R}^{(T_{\mathrm{ini}}-1)m\times (T_{\mathrm{ini}}-1)m}}
\Delta U_p,
\\
\Delta \tilde{U}_f 
&= 
\underbrace{
\begin{bmatrix}
I_{m \times m} & 0 & 0 & \cdots & 0 \\
I_{m \times m} & I_{m \times m} & 0 & \cdots & 0 \\
\vdots & \vdots & \ddots & \ddots & \vdots \\
I_{m \times m} & I_{m \times m} & I_{m \times m} & \cdots & I_{m \times m}
\end{bmatrix}}
_{Q_{U_f}\in \mathbb{R}^{Nm\times Nm}}
\Delta U_f,
\\
\Delta \tilde{Y}_p 
&= 
\underbrace{
\begin{bmatrix}
I_{p \times p} & 0 & 0 & \cdots & 0 \\
I_{p \times p} & I_{p \times p} & 0 & \cdots & 0 \\
\vdots & \vdots & \ddots & \ddots & \vdots \\
I_{p \times p} & I_{p \times p} & I_{p \times p} & \cdots & I_{p \times p}
\end{bmatrix}}
_{Q_{Y_p}\in \mathbb{R}^{(T_{\mathrm{ini}}-1)p\times (T_{\mathrm{ini}}-1)p}} 
\Delta Y_p,
\\
\Delta \tilde{Y}_f 
&= 
\underbrace{
\begin{bmatrix}
I_{p \times p} & 0 & 0 & \cdots & 0 \\
I_{p \times p} & I_{p \times p} & 0 & \cdots & 0 \\
\vdots & \vdots & \ddots & \ddots & \vdots \\
I_{p \times p} & I_{p \times p} & I_{p \times p} & \cdots & I_{p \times p}
\end{bmatrix}}
_{Q_{Y_f}\in \mathbb{R}^{Np\times Np}}
\Delta Y_f,
\\
\Delta \tilde{u}_{\mathrm{ini}}
&= Q_{U_p} \cdot \Delta u_{\mathrm{ini}},
\\
\Delta \tilde{u}
&= Q_{U_f} \cdot \Delta u,
\\
\Delta \tilde{y}_{\mathrm{ini}}
&= Q_{Y_p} \cdot \Delta y_{\mathrm{ini}},
\\
\Delta \tilde{y}
&= Q_{Y_f} \cdot \Delta y,
\end{aligned}
\end{equation}
where $I_{m\times m}$ and $I_{p\times p}$ represent the $m$-by-$m$ and $p$-by-$p$ identity matrices, respectively.
In~\eqref{equ:Delta_tilde_U_Y}, the matrices $Q_{U_p}$, $Q_{U_f}$, $Q_{Y_p}$, and $Q_{Y_f}$ are all square and invertible, and thus the non-parametric representation in~\eqref{equ:Delta_U_Y} can be reformulated as 
\begin{equation}\label{eq:final_eq}
\begin{bmatrix}
\Delta \tilde{U}_p \\
\Delta \tilde{U}_f \\
\Delta \tilde{Y}_p \\
\Delta \tilde{Y}_f
\end{bmatrix}
g =
\begin{bmatrix}
\Delta \tilde{u}_{\mathrm{ini}} \\
\Delta \tilde{u} \\
\Delta \tilde{y}_{\mathrm{ini}} \\
\Delta \tilde{y}
\end{bmatrix}.
\end{equation}

Building on the regularized DeePC~\eqref{eq:deepc_formulation} and the non-parametric representation~\eqref{eq:final_eq}, we formulate the velocity-form DeePC at time instant $t$ as follows:
\begin{equation}\label{eq:vf_deepc_formulation}
\begin{aligned}
\min_{g, u, y, \sigma_y} \quad &
\| y - y^r \|_Q^2 + \| u \|_R^2 + \lambda_g \| g \|_2^2 + \lambda_y \| \sigma_y \|_2^2 \\
\text{subject to} \quad &
\begin{bmatrix}
\Delta \tilde{U}_p \\
\Delta \tilde{U}_f \\
\Delta \tilde{Y}_p \\
\Delta \tilde{Y}_f
\end{bmatrix}
g =
\begin{bmatrix}
\Delta \tilde{u}_{\mathrm{ini}} \\
\Delta \tilde{u} \\
\Delta \tilde{y}_{\mathrm{ini}} \\
\Delta \tilde{y}
\end{bmatrix}
+
\begin{bmatrix}
0 \\
0 \\
\sigma_y \\
0
\end{bmatrix}, \\
& y = \mathbf{1}_N \otimes y_{t-1} + \Delta \tilde{y}, \\
& u = \mathbf{1}_N \otimes u_{t-1} + \Delta \tilde{u}, \\
& y \in \mathcal{Y}, \quad u \in \mathcal{U},
\end{aligned}
\end{equation}
where $\mathbf{1}_N$ is an $N$-dimensional vector of ones, and $\otimes$ denotes the Kronecker product.


\eqref{eq:final_eq} can be interpreted as the velocity-form non-parametric representation of system~\eqref{eq:linear_syst_dist}. When the disturbances in~\eqref{eq:linear_syst_dist} are constant, this representation can effectively eliminate their influence, thereby enabling offset-free prediction. By combining the regularized DeePC formulation in Section~\ref{sub_deepc} with the offset-free property of the velocity-form non-parametric representation, the proposed formulation~\eqref{eq:vf_deepc_formulation} is expected to deliver improved control performance in the presence of additional payloads. Although payloads are not strictly constant during dynamic motion, approximating them as constant disturbances yields better performance than neglecting them entirely. Experimental results in Section~\ref{sec_experiments} validate this expectation, demonstrating that velocity-form DeePC achieves superior performance under unknown payload conditions compared to standard DeePC.

An offset-free DeePC formulation was proposed in~\cite{Lazar2022}. Their approach adopts the incremental form only for the control input, rendering it primarily effective in mitigating state disturbances. In contrast, our formulation employs the incremental form for both inputs and outputs, thereby enhancing robustness against output disturbances as well. Moreover, \cite{Lazar2022} primarily aimed to elucidate the relationships among offset-free MPC, subspace predictive control, and offset-free DeePC, and our work is centered on developing a new DeePC formulation tailored to improve the control performance of soft robots under unknown payloads.

\subsection{Mosaic Hankel Matrix and Dimension Reduction}\label{matrices_ex_re} 
In many cases, it is not possible to obtain long trajectories.  
The system might be unstable, or a single trajectory over that period may be insufficiently excited to accurately capture the true system dynamics.  
Therefore, combining multiple short trajectories can be beneficial~\cite{Waarde2020}.  
Mosaic Hankel matrices are utilized in such situations.

\begin{definition} [\cite{Waarde2020}]\label{def:collective_pe}
Consider signal sequences $\omega^{i}_{[0,T_i-1]}$ for $i=1, 2, \ldots, \kappa$, where $\kappa$ denotes the number of datasets and $T_{i}$ are positive integers.
Define a positive integer $\psi$ such that $\psi \leq T_i$ for all $i$. 
The sequences $\omega^{i}_{[0,T_i-1]}$ for $i = 1, 2, \ldots, \kappa$ are called \textit{collectively persistently exciting} of order $\psi$ if the mosaic-Hankel matrix
\[
\begin{bmatrix}
\mathcal{H}_\psi\left(\omega^1_{[0,T_1-1]}\right) &
\mathcal{H}_\psi\left(\omega^2_{[0,T_2-1]}\right) &
\cdots &
\mathcal{H}_\psi\left(\omega^\kappa_{[0,T_\kappa-1]}\right)
\end{bmatrix}
\]
has full row rank.
\end{definition}

Suppose $\kappa$ sets of input and output trajectories $u^{i}_{[0,T_{i}-1]}$ and $y^{i}_{[0,T_{i}-1]}$, each of length $T_{i}$, are collected. The corresponding incremental trajectories $\Delta u^{i}_{[0,T_{i}-2]}$ and $\Delta y^{i}_{[0,T_{i}-2]}$ are constructed by following~\eqref{equ:Delta}. Then, according to Definition~\ref{def:collective_pe}, all Hankel matrices $\mathcal{H}_{L-1}(\Delta u_{[0, T_{i}-2]}^{i})$ and $\mathcal{H}_{L-1}(\Delta y_{[0, T_{i}-2]}^{i})$ are concatenated to form the mosaic Hankel matrix, as follows:
\begin{equation}\label{eq:mosiac}
\mathcal{H}_{L-1}^{\Delta}
:=
\begin{bmatrix}
\mathcal{H}_{L-1} (\Delta u^1_{[0,T_{1}-2]}) &
\cdots &
\mathcal{H}_{L-1} (\Delta u^{\kappa}_{[0,T_{\kappa}-2]}) \\
\mathcal{H}_{L-1}(\Delta y^1_{[0,T_{1}-2]}) &
\cdots &
\mathcal{H}_{L-1}(\Delta y^{\kappa}_{[0,T_{\kappa}-2]})
\end{bmatrix}.
\end{equation}
As shown in Theorem 2 of~\cite{Waarde2020}, if the incremental input sequences $\Delta u^{i}_{[0,T_{i}-2]}$ are collectively persistently exciting of order $n+L-1$, then the mosaic-Hankel matrix $\mathcal{H}_{L-1}^{\Delta}$ in~\ref{eq:mosiac} enables a non-parametric representation of the system. Thus, in the velocity-form DeePC~\eqref{eq:vf_deepc_formulation}, one can use $\mathcal{H}_{L-1}^{\Delta}$ to replace the matrix $\begin{bmatrix}
\Delta \tilde{U}_p^{\top} &
\Delta \tilde{U}_f^{\top} &
\Delta \tilde{Y}_p^{\top} &
\Delta \tilde{Y}_f^{\top}
\end{bmatrix}^{\top}$. 
This completes the matrix extension step. 

To ensure collective persistent excitation of order $n+L-1$, the total data length should satisfy $\sum_{i=1}^{\kappa}T_{i} \ge (m+\kappa)(n+L-1)-\kappa$~\cite{Waarde2020}. 
In practical applications, one would need to collect a sufficiently large number of data points for each dataset. This will result in a very large value of $\sum_{i=1}^{\kappa}T_{i}$, thus making the velocity-form DeePC problem large-scale and nontrivial to solve efficiently.  
To address this issue, the singular value decomposition (SVD) technique~\cite{Zhang2023} can be applied to reduce the dimension of the mosaic-Hankel matrix~\eqref{eq:mosiac}.  
Specifically, we perform SVD on $\mathcal{H}_{L-1}^{\Delta}$, which can be expressed as:
\begin{equation} \label{eq:svd}
\mathcal{H}_{L-1}^{\Delta}
=
\underbrace{\begin{bmatrix}
W_1 & W_2
\end{bmatrix}}_{W}
\underbrace{\begin{bmatrix}
\Sigma_1 & 0 \\
0 & \Sigma_2
\end{bmatrix}}_{\Sigma}
\underbrace{\begin{bmatrix}
V_1 & V_2
\end{bmatrix}^\top}_{V^\top},
\end{equation}
where $W \in \mathbb{R}^{q_1 \times q_1}$ and $V \in \mathbb{R}^{q_2 \times q_2}$ are orthogonal matrices, with $q_1 = (m+p)\times(L-1)  $ and $q_2 = \sum_{i=1}^{\kappa} (T_{i} - L + 1)$. The matrix $\Sigma \in \mathbb{R}^{q_1 \times q_2}$ is a rectangular diagonal matrix with non-negative singular values on the diagonal, arranged in descending order.  

Let $\Sigma_1 \in \mathbb{R}^{r \times r}$ ($r \leq \min\{q_1, q_2\}$) contain the top $r$ nonzero singular values. Then $\Sigma_2$, $W_1$, $W_2$, $V_1$, and $V_2$ are defined with appropriate dimensions. The new data matrix $\tilde{\mathcal{H}}^\Delta_L \in \mathbb{R}^{q_1 \times r}$ is constructed via
\begin{equation} \label{eq:svd2}
\tilde{\mathcal{H}}^\Delta_{L-1} = \mathcal{H}^\Delta_{L-1} V_1 = W_1 \Sigma_1.
\end{equation}
This yields a more compact matrix representation that preserves the essential information of the system dynamics. 

After the extension and reduction techniques, the optimization problem is updated to the following representation:
\begin{equation}\label{eq:vf_ext_deepc_formulation}
\begin{aligned}
\min_{\tilde{g}, u, y, \sigma_y} \quad &
\| y - y^{r} \|_Q^2 + \| u \|_R^2 + \lambda_g \| g \|_2^2 + \lambda_y \| \sigma_y \|_2^2 \\
\text{subject to} \quad &
\tilde{\mathcal{H}}^\Delta_L
\tilde{g} =
\begin{bmatrix}
\Delta \tilde{u}_{\text{ini}} \\
\Delta \tilde{u} \\
\Delta \tilde{y}_{\text{ini}} \\
\Delta \tilde{y}
\end{bmatrix}
+
\begin{bmatrix}
0 \\
0 \\
\sigma_y \\
0
\end{bmatrix}, \\
& y = \mathbf{1}_N \otimes y_{t-1} + \Delta \tilde{y}, \\
& u = \mathbf{1}_N \otimes u_{t-1} + \Delta \tilde{u}, \\
& y \in \mathcal{Y}, \quad u \in \mathcal{U}.
\end{aligned}
\end{equation}
In this formulation, $\tilde{g}$ has a reduced dimension of $r$, while the dimension of the optimization variable will be $\sum_{i=1}^{\kappa}(T_{i}-L+1)$ when the original mosaic-Hankel matrix $\mathcal{H}^{\Delta}_{L-1}$ is utilized. This reduction effectively decreases the size of the data matrix and lowers the computational time. 

The implementation of the velocity-form DeePC framework is summarized in Algorithm~\ref{algo:vdeepc}. 
The process begins with constructing the mosaic Hankel matrix, followed by SVD-based dimension reduction. 
The reduced matrices are then incorporated into the optimization problem to compute the optimal control sequence. The first control input is applied, and the process repeats in a receding-horizon manner.

\begin{algorithm}
  \caption{Velocity-Form DeePC Algorithm for Soft Robot} \label{algo:vdeepc}
  \begin{algorithmic}[1] %
    \State \textbf{Input:} Final step $t_{\mathrm{end}}$, pre-collected pressure input $u^{i}_{[0, T_{i}-1]}$, soft robot position output $y^{i}_{[0, T_{i}-1]}$, $i=1,2,\ldots,\kappa$.
    \State Compute the incremental input and output trajectories $\Delta u^{i}_{[0,T_{i}-2]}$ and $\Delta y^{i}_{[0,T_{i}-2]}$.
    \State Construct mosaic-Hankel matrix based on~\eqref{eq:mosiac}. 
    \State Perform SVD on mosaic-Hankel matrix via~\eqref{eq:svd} and then compute the new data matrix via~\eqref{eq:svd2}.
    \State For $t<T_{\mathrm{ini}}$, initialize $u_{\mathrm{ini}}$ with 0 and $y_{\mathrm{ini}}$ with real-time position measurements.
    \While {$T_{\mathrm{ini}}\le t \le t_{\mathrm{end}}$} \label{algo:line:startwhile}
      \State Solve the optimization problem \eqref{eq:vf_ext_deepc_formulation} to obtain optimal pressure control sequence $u^{*}$.
      \State  Send the first step optimal pressure control $u^{*}_0$ to low-level pneumatic controller.
      \State  Measure the soft robot position, and update $u_{\mathrm{ini}}$ and $y_{\mathrm{ini}}$ to the $T_{\mathrm{ini}}$ most recent input/output measurements.
      \State  Set $t$ to $t+1$.
    \EndWhile \label{algo:line:endwhile}
  \end{algorithmic}
\end{algorithm}

    \begin{figure}[!h]
		\centering
		\includegraphics[width=1\linewidth]{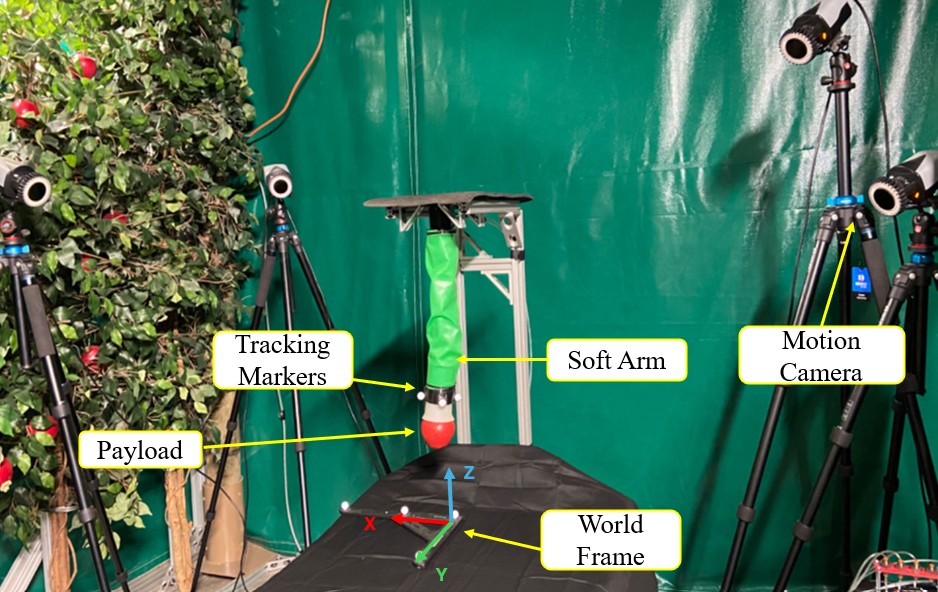}
		\caption{The experimental setup of the soft robot control system. The soft robot is surrounded by a Qualisys motion capture system equipped with eight cameras.}\label{fig_softarm_setup}
	\end{figure}

\section{Experiments} \label{sec_experiments}
In this section, we present the experimental validation of the velocity-form DeePC (VDeePC) in comparison with the standard DeePC. The evaluation is conducted under both unloaded and loaded conditions, where apples of varying weights are used as unknown payloads.

\subsection{Experimental Setup} \label{sub_setup}
The experimental setup of the soft robot system is shown in Fig.~\ref{fig_softarm_setup}. The soft arm is suspended vertically downward on an aluminum frame, with air supply lines connected to the robot and routed behind the frame. The setup is surrounded by eight Qualisys motion capture cameras.  

The DeePC algorithm is implemented in MATLAB on a host computer, and its control commands are transmitted to a low-level Arduino-based embedded controller via an Ethernet cable. The low-level controller drives an open-source pneumatic control board equipped with valves and pressure sensors; details of the board design and setup can be found in \cite{Wang2025,Holland2014}. Communication between the host and the embedded controller is handled through the ROS publisher–subscriber interface. The low-level controller regulates fast-switching on-off electronic solenoid valves (SMC VQ110U-5M) for accurate pressure control. These valves modulate airflow from the external air supply into the soft robot system, while pressure sensors provide feedback to achieve closed-loop control.  

A suction gripper is mounted on the end effector to facilitate payload handling. For payload experiments, an apple is manually attached to the gripper. The suction gripper is included solely to demonstrate the feasibility of payload-carrying strategies and is not the primary focus of this study. Future work will consider the implementation of more realistic gripping mechanisms, such as vacuum-driven systems.

        \subsection{Data Collection} \label{sub_collect}

To collect persistently exciting data for DeePC, we generate random input sequences in the form of pressure control commands, similar to~\cite{Bruder2021}. After collecting several different trajectories, we construct Hankel matrices for each short trajectory and then follow the procedures described in Section~\ref{matrices_ex_re} to perform matrix extension and reduction. For the extension step, we concatenate the individual Hankel matrices, stacking four separate datasets—each with a duration of 601 time steps—to form a rich and informative Hankel matrix suitable for behavioral system representation. Both the control inputs and motion capture data are sampled at a rate of 10~Hz. For the reduction step, we apply SVD and select $r = 400$. In this way, the original matrix with a column size of $(T - L + 1)\kappa = 2248$, where $m = 1$, $p = 2$, $T_{\text{ini}} = 20$, $N = 20$, and $L = T_{\text{ini}} + N$, is reduced to a column size of $400$.

The control command pressure is constrained within the range $[0, 80]$~psi. The workspace for the robot end-effector is bounded as  $[240, 350]$~mm along the $z$-direction and $[30, 160]$~mm along the $x$-direction based on the world frame. We evaluate the control performance in multiple experiments. First, we perform setpoint tracking tasks using both the standard DeePC (specifically, the regularized DeePC formulation) and the velocity-form DeePC, under both unloaded and loaded conditions. Next, we assess robustness by attaching apples of varying weights to the robot tip. Finally, both control schemes are evaluated along a long trajectory involving abrupt changes in the reference setpoint. 
 
  \begin{figure}[!h]
		\centering
            \vspace{-45pt}
		\includegraphics[width=1\linewidth]{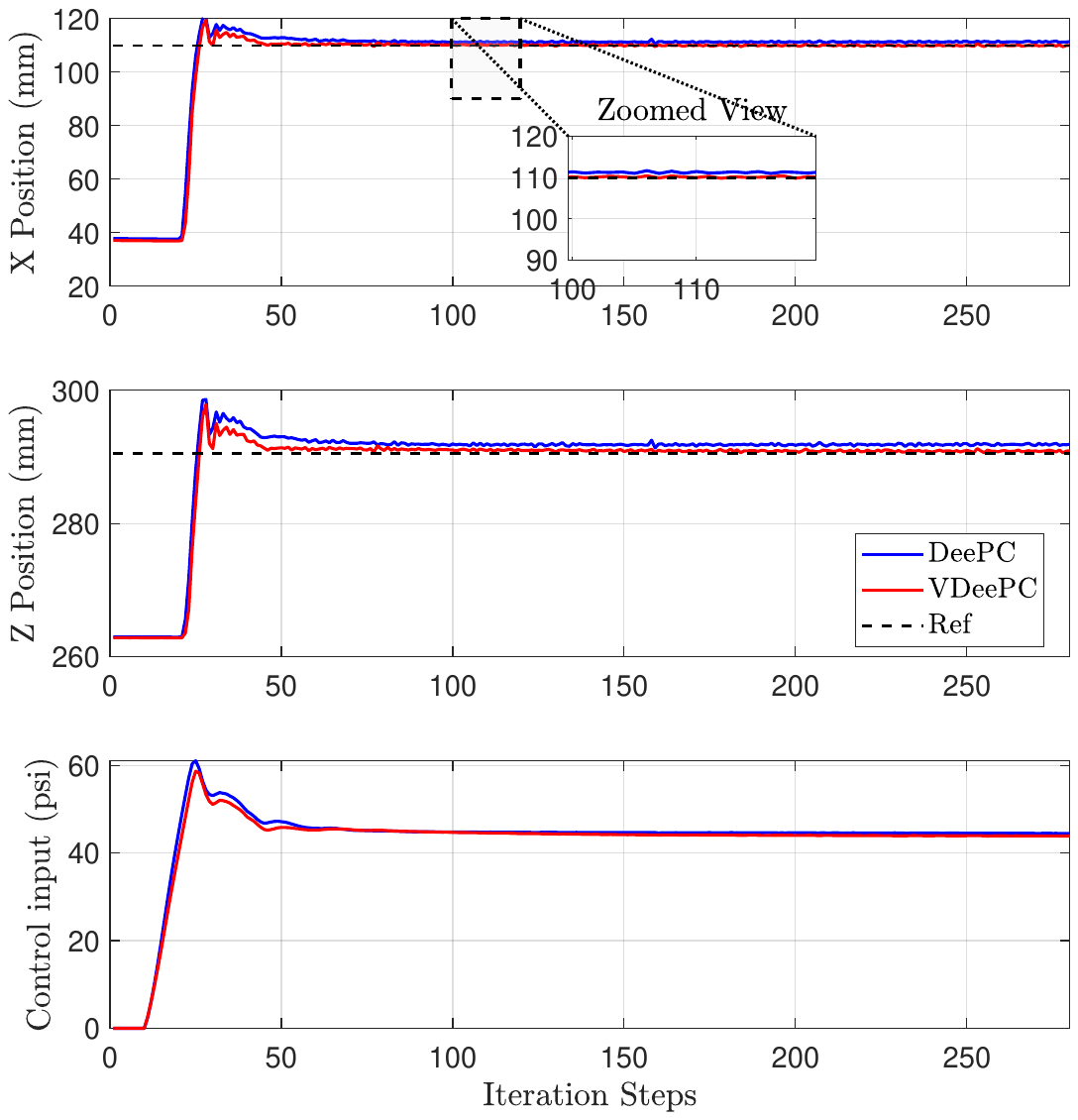}
            \vspace{-55pt}
		\caption{The setpoint tracking results for the DeePC vs VDeePC without load.}\label{fig_setpoint}
        \vspace{15pt}
	\end{figure}

  \begin{figure}[!h]
		\centering
            \vspace{-45pt}
		\includegraphics[width=1\linewidth]{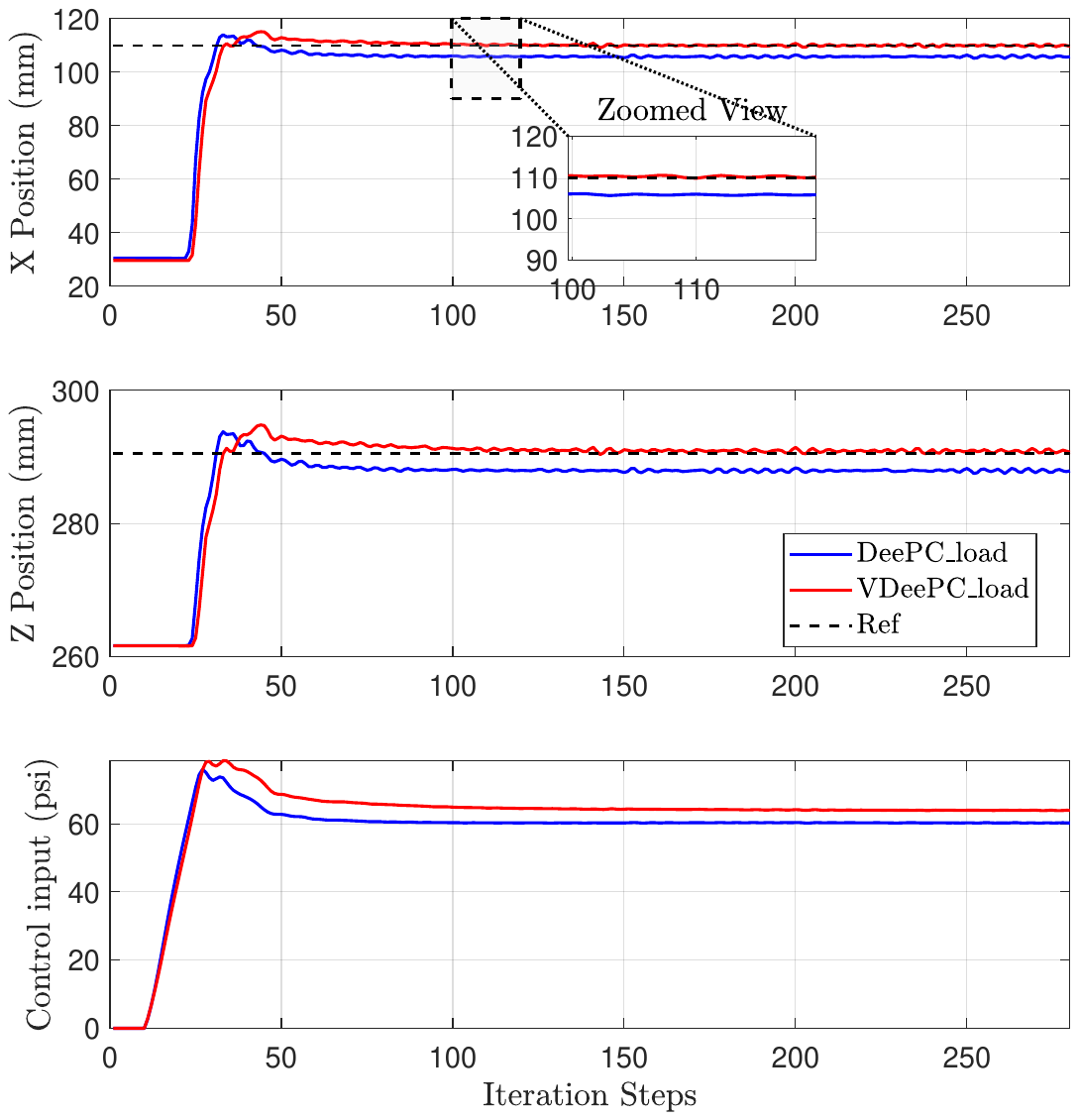}
            \vspace{-55pt}
		\caption{The setpoint tracking results for the DeePC vs VDeePC with load.}\label{fig_setpoint_load}
        \vspace{15pt}
	\end{figure}

    \subsection{Setpoint Control Evaluation} \label{sub_setpoint}

In this section, we evaluate the setpoint regulation performance of both the standard DeePC algorithm and its velocity-form variant. The hyperparameters are selected as follows: $Q = 10^{6}$, $R = 10^{-6}$, $\lambda_{g} = 10^{3}$, and $\lambda_{y} = 10^{5}$. A feasible reference point \((x_{\mathrm{ref}} = 109.9, \; z_{\mathrm{ref}} = 290.5)\) in the robot's task space is selected, and both controllers are tasked with tracking this reference. As shown in Fig.~\ref{fig_setpoint}, both algorithms demonstrate satisfactory tracking performance under no-load conditions, with control commands remaining within comparable and acceptable ranges. Notably, the VDeePC exhibits improved tracking performance, with a root mean square error (RMSE) in Euclidean distance of 0.47~mm, compared to 1.89~mm for DeePC.

Next, we introduce an external perturbation by attaching a small apple to the tip of the soft robotic manipulator. This added mass changes the system dynamics in a way that was not present during the offline data collection stage. As shown in Fig.~\ref{fig_setpoint_load}, the original DeePC controller, which performed well under nominal conditions, exhibits noticeable undershoot in both the $x$ and $z$ directions, indicating sensitivity to the model mismatch introduced by the added payload. In contrast, VDeePC continues to track the setpoint with minimal degradation in performance. The RMSE is 4.90~mm for DeePC and 0.53~mm for VDeePC. The steady-state error is reduced by 89$\%$, demonstrating that the velocity-form formulation offers improved performance against unmodeled disturbances or plant variations.

\subsection{Tracking Performance Under Varying Weights} \label{sub_weights}

\begin{figure}[!h]
		\centering
		\includegraphics[width=1\linewidth]{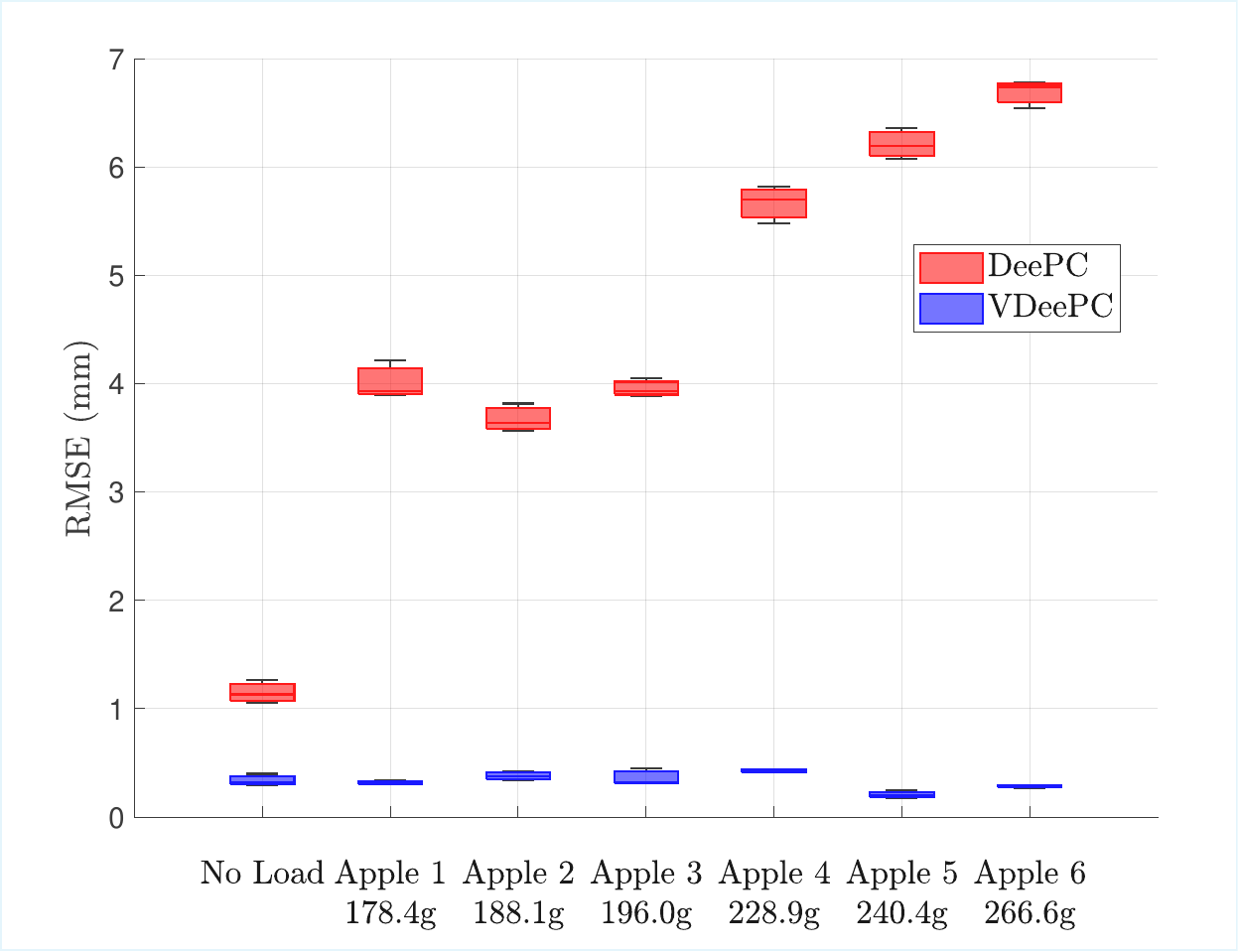}
		\caption{ Boxplot results of tracking errors under no-load and load conditions with six apples of different weights, comparing DeePC and VDeePC.}\label{fig_apple}
	\end{figure}

To further investigate robustness, we evaluate both algorithms in tracking the same reference points as in the previous section, but this time under a range of external payloads. Specifically, we attach apples of different known weights to the robot and assess how tracking accuracy is affected. Fig.~\ref{fig_apple} presents tracking results across these scenarios. Each test was repeated three times to ensure consistency. 

The test weights used in this study are 178.4~g, 188.05~g, 196.02~g, 228.96~g, 240.43~g, and 266.59~g. When no apple is attached, both controllers achieve low tracking errors, with the VDeePC slightly outperforming the standard DeePC. The variance of the error (as shown by the box plot spread) remains small for both methods, indicating good repeatability and robustness. As the payload increases, the DeePC tracking error grows, whereas the VDeePC consistently yields lower median RMSE values across all payloads, remaining below 0.42~mm in all cases. In contrast, the DeePC median tracking error ranges from 1.13~mm to 6.74~mm. The results suggest that the VDeePC offers superior disturbance rejection and better performance under payload variation, even though these disturbances were not explicitly included in the training dataset.

 \begin{figure}[!h]
		\centering
        \vspace{-45pt}
		\includegraphics[width=1\linewidth]{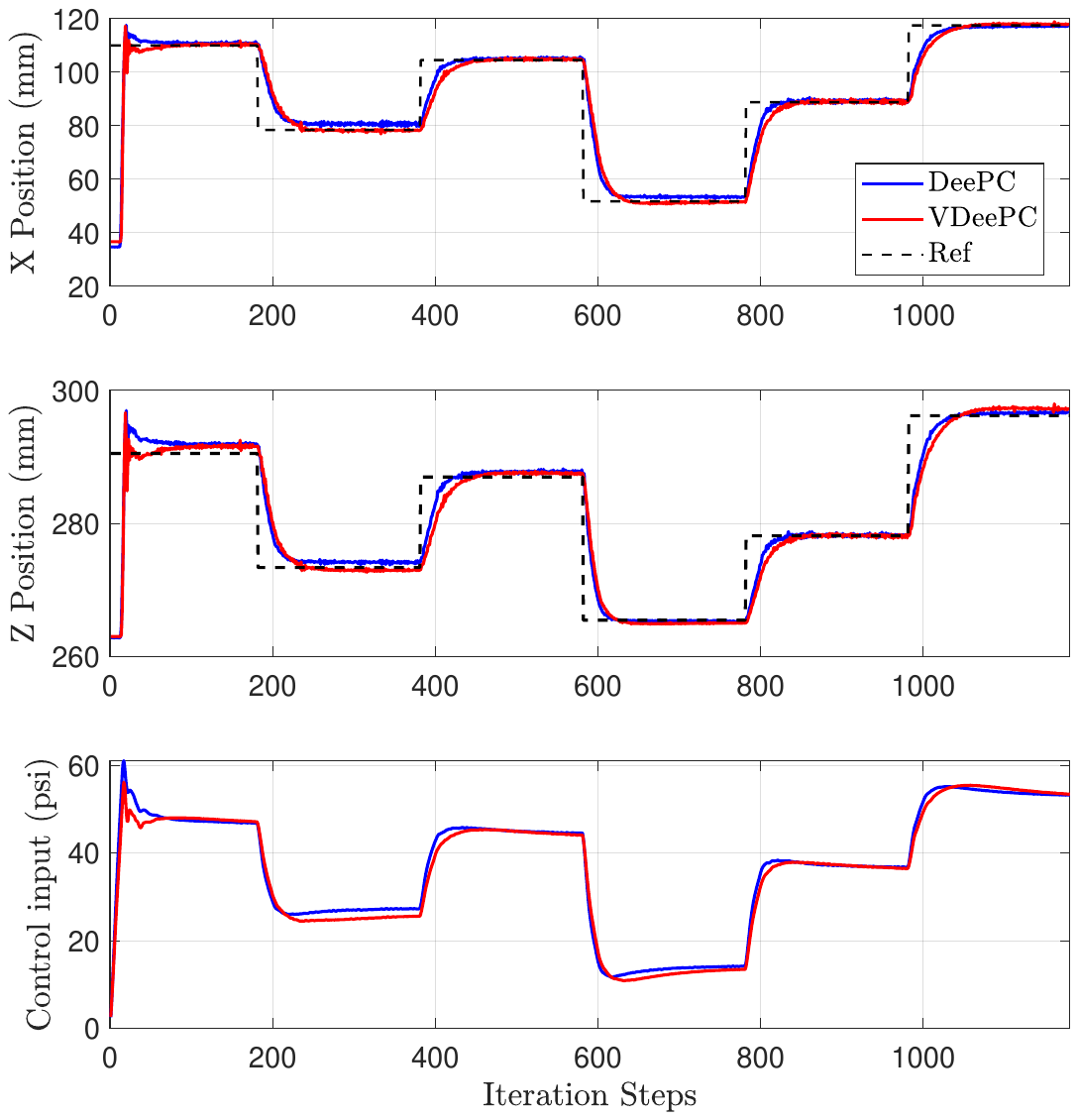}
        \vspace{-55pt}
		\caption{The tracking results for the DeePC vs VDeePC without load.}\label{fig_ramp}
        \vspace{+15pt}

	\end{figure}

  \begin{figure}[!h]
		\centering
        \vspace{-45pt}
		\includegraphics[width=1\linewidth]{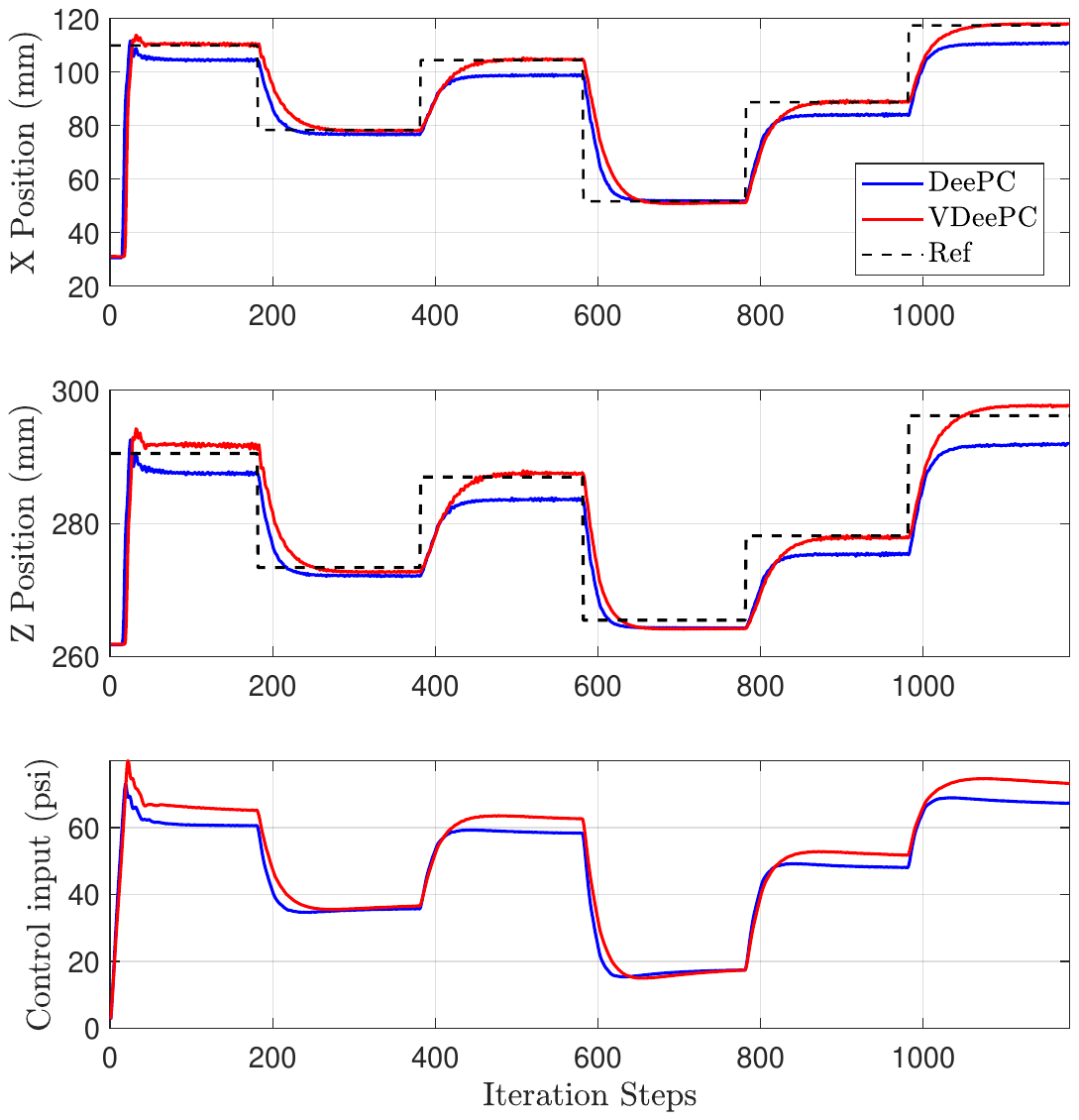}
        \vspace{-55pt}
		\caption{The tracking results for the DeePC vs VDeePC with load.}\label{fig_ramp_load}
        \vspace{+15pt}
	\end{figure}

\subsection{Long Trajectory Tracking with Multiple Reference Changes} \label{sub_long}

In this section, we evaluate the performance and adaptability of the standard DeePC and VDeePC algorithms when tracking a long trajectory containing multiple changes in the desired reference.

Initially, we conduct the experiment without any external payload. As shown in Fig.~\ref{fig_ramp}, both controllers demonstrate good performance in following the  reference trajectory. The responses from both DeePC and VDeePC exhibit fast convergence with minimal overshoot, and their control inputs remain within acceptable bounds throughout the trajectory. Both methods achieve accurate and stable reference tracking.

Next, we repeat the same trajectory tracking task with an added payload — a 178~g apple affixed to the end effector of the soft robot. Under the loaded condition, we observe a significant difference in the performance of the two controllers, as shown in Fig.~\ref{fig_ramp_load}. The standard DeePC algorithm begins to show noticeable steady-state errors, especially in trajectory segments that require the robot to lift the apple. In particular, it consistently fails to reach the desired positions in the $z$ direction due to the unmodeled gravitational effect of the added mass. 

In contrast, the VDeePC controller continues to perform well, closely tracking the reference throughout the entire trajectory despite the additional payload, and demonstrates significantly improved steady-state performance. The average RMSE across all steady-state regions is 6.41~mm for DeePC and 3.07~mm for VDeePC. This improvement is due to the velocity-form formulation, which leverages incremental system behavior and naturally incorporates integral-like action. As a result, robustness to model mismatch and external disturbances is improved.


	\section{Conclusion} \label{sec_conclusion}
 
This paper presented a novel velocity-form DeePC approach for controlling a planar soft robot subject to unknown payloads. We began by reviewing the preliminaries of DeePC, and then introduced the velocity-form DeePC strategy. To validate the proposed method, we conducted a series of experiments comparing standard DeePC and velocity-form DeePC under both unloaded and loaded conditions. Notably, the velocity-form DeePC demonstrated superior performance in the presence of unmodeled external loads, achieving accurate tracking without requiring disturbance modeling or re-identification. This is particularly significant given that no prior data collection was performed under the loaded conditions. Overall, the experimental results confirmed the effectiveness and robustness of the proposed velocity-form DeePC approach, highlighting its potential for soft robot applications in dynamic and uncertain environments. The current work focuses on a single-segment planar soft robot under load. Future work will extend the proposed control framework to soft robots with additional degrees of freedom and multiple segments.

	\balance	
	\bibliographystyle{IEEEtran}
	\bibliography{IEEEabrv,reference}

\begin{thebibliography}{10}
\providecommand{\url}[1]{#1}
\csname url@samestyle\endcsname
\providecommand{\newblock}{\relax}
\providecommand{\bibinfo}[2]{#2}
\providecommand{\BIBentrySTDinterwordspacing}{\spaceskip=0pt\relax}
\providecommand{\BIBentryALTinterwordstretchfactor}{4}
\providecommand{\BIBentryALTinterwordspacing}{\spaceskip=\fontdimen2\font plus
\BIBentryALTinterwordstretchfactor\fontdimen3\font minus \fontdimen4\font\relax}
\providecommand{\BIBforeignlanguage}[2]{{%
\expandafter\ifx\csname l@#1\endcsname\relax
\typeout{** WARNING: IEEEtran.bst: No hyphenation pattern has been}%
\typeout{** loaded for the language `#1'. Using the pattern for}%
\typeout{** the default language instead.}%
\else
\language=\csname l@#1\endcsname
\fi
#2}}
\providecommand{\BIBdecl}{\relax}
\BIBdecl

\bibitem{Lee2024}
K.~Lee, K.~Bayarsaikhan, G.~Aguilar, J.~Realmuto, and J.~Sheng, ``Design and characterization of soft fabric omnidirectional bending actuators,'' \emph{Actuators}, vol.~13, no.~3, 2024.

\bibitem{Liu2024}
T.~Liu, T.~Abrar, and J.~Realmuto, ``Modular and reconfigurable body mounted soft robots,'' in \emph{2024 IEEE 7th International Conference on Soft Robotics (RoboSoft)}, 2024, pp. 145--150.

\bibitem{Xavier2022}
M.~S. Xavier, C.~D. Tawk, A.~Zolfagharian, J.~Pinskier, D.~Howard, T.~Young, J.~Lai, S.~M. Harrison, Y.~K. Yong, M.~Bodaghi, and A.~J. Fleming, ``Soft pneumatic actuators: A review of design, fabrication, modeling, sensing, control and applications,'' \emph{IEEE Access}, vol.~10, pp. 59\,442--59\,485, 2022.

\bibitem{Armanini2023}
C.~Armanini, F.~Boyer, A.~T. Mathew, C.~Duriez, and F.~Renda, ``Soft robots modeling: A structured overview,'' \emph{IEEE Transactions on Robotics}, vol.~39, no.~3, pp. 1728--1748, 2023.

\bibitem{Wang2021}
J.~Wang and A.~Chortos, ``Control strategies for soft robot systems,'' \emph{Advanced Intelligent Systems}, vol.~4, no.~5, p. 2100165.

\bibitem{ScheggDuriez2022}
P.~Schegg and C.~Duriez, ``Review on generic methods for mechanical modeling, simulation and control of soft robots,'' \emph{PLoS ONE}, vol.~17, no.~1, p. e0251059, 2022.

\bibitem{Duriez2013}
C.~Duriez, ``Control of elastic soft robots based on real-time finite element method,'' in \emph{2013 IEEE International Conference on Robotics and Automation}, 2013, pp. 3982--3987.

\bibitem{Jones2009}
B.~A. Jones, R.~L. Gray, and K.~Turlapati, ``Three dimensional statics for continuum robotics,'' in \emph{2009 IEEE/RSJ International Conference on Intelligent Robots and Systems}, 2009, pp. 2659--2664.

\bibitem{Webster2010}
R.~J. Webster and B.~A. Jones, ``Design and kinematic modeling of constant curvature continuum robots: A review,'' vol.~29, no.~13, 2010.

\bibitem{Hannan2003}
M.~W. Hannan and I.~D. Walker, ``Kinematics and the implementation of an elephant's trunk manipulator and other continuum style robots,'' \emph{Journal of Robotic Systems}, vol.~20, no.~2, pp. 45--63, 2003.

\bibitem{Katzschmann2019}
R.~K. Katzschmann, C.~D. Santina, Y.~Toshimitsu, A.~Bicchi, and D.~Rus, ``Dynamic motion control of multi-segment soft robots using piecewise constant curvature matched with an augmented rigid body model,'' in \emph{2019 2nd IEEE International Conference on Soft Robotics (RoboSoft)}, 2019, pp. 454--461.

\bibitem{Falkenhahn2014}
V.~Falkenhahn, T.~Mahl, A.~Hildebrandt, R.~Neumann, and O.~Sawodny, ``Dynamic modeling of constant curvature continuum robots using the euler-lagrange formalism,'' in \emph{2014 IEEE/RSJ International Conference on Intelligent Robots and Systems}, 2014, pp. 2428--2433.

\bibitem{Cheng2024}
H.~Cheng, L.~Chen, B.~Fang, J.~Zhang, and J.~Hong, ``Switching control for a soft rehabilitation glove with pneumatic bellows actuators,'' \emph{IEEE Robotics and Automation Letters}, vol.~9, no.~6, pp. 4966--4973, 2024.

\bibitem{Thuruthel2019}
T.~G. Thuruthel, E.~Falotico, F.~Renda, and C.~Laschi, ``Model-based reinforcement learning for closed-loop dynamic control of soft robotic manipulators,'' \emph{IEEE Transactions on Robotics}, vol.~35, no.~1, pp. 124--134, 2019.

\bibitem{Satheeshbabu2019}
S.~Satheeshbabu, N.~K. Uppalapati, G.~Chowdhary, and G.~Krishnan, ``Open loop position control of soft continuum arm using deep reinforcement learning,'' in \emph{2019 International Conference on Robotics and Automation (ICRA)}, 2019, pp. 5133--5139.

\bibitem{Bruder2021TRO}
D.~Bruder, X.~Fu, R.~B. Gillespie, C.~D. Remy, and R.~Vasudevan, ``Data-driven control of soft robots using koopman operator theory,'' \emph{IEEE Transactions on Robotics}, vol.~37, no.~3, pp. 948--961, 2021.

\bibitem{Wang2023}
J.~Wang, B.~Xu, J.~Lai, Y.~Wang, C.~Hu, H.~Li, and A.~Song, ``An improved koopman-mpc framework for data-driven modeling and control of soft actuators,'' \emph{IEEE Robotics and Automation Letters}, vol.~8, no.~2, pp. 616--623, 2023.

\bibitem{WangJiajin2025}
J.~Wang, B.~Xu, J.~Liu, Z.~Zhao, W.~Peng, and A.~Song, ``Robust koopman-mpc approach with high-order disturbance observer for control of pneumatic soft bending actuators under external loads,'' \emph{IEEE/ASME Transactions on Mechatronics}, pp. 1--12, 2025.

\bibitem{Wang2025}
H.~Wang, K.~Zhang, K.~Lee, Y.~Mei, K.~Zhu, V.~Srivastava, J.~Sheng, and Z.~Li, ``Mechanical design and data-enabled predictive control of a planar soft robot,'' \emph{IEEE Robotics and Automation Letters}, vol.~9, no.~9, pp. 7923--7930, 2024.

\bibitem{Cao2021}
Y.~Cao, J.~Huang, H.~Ru, W.~Chen, and C.-H. Xiong, ``A visual servo-based predictive control with echo state gaussian process for soft bending actuator,'' \emph{IEEE/ASME Transactions on Mechatronics}, vol.~26, no.~1, pp. 574--585, 2021.

\bibitem{Ji2025}
G.~Ji, Q.~Gao, Y.~Xiao, and Z.~Sun, ``Efficient real2sim2real of continuum robots using deep reinforcement learning with koopman operator,'' \emph{IEEE Transactions on Industrial Electronics}, vol.~72, no.~8, pp. 8333--8343, 2025.

\bibitem{Bruder2021}
D.~Bruder, X.~Fu, R.~B. Gillespie, C.~D. Remy, and R.~Vasudevan, ``Koopman-based control of a soft continuum manipulator under variable loading conditions,'' \emph{IEEE Robotics and Automation Letters}, vol.~6, no.~4, pp. 6852--6859, 2021.

\bibitem{Pannocchia2015}
G.~Pannocchia, ``Offset-free tracking mpc: A tutorial review and comparison of different formulations,'' in \emph{2015 European Control Conference (ECC)}, 2015, pp. 527--532.

\bibitem{CHEN2022102871}
J.~Chen, Y.~Dang, and J.~Han, ``Offset-free model predictive control of a soft manipulator using the koopman operator,'' \emph{Mechatronics}, vol.~86, p. 102871, 2022.

\bibitem{WILLEMS2005325}
J.~C. Willems, P.~Rapisarda, I.~Markovsky, and B.~L. {De Moor}, ``A note on persistency of excitation,'' \emph{Systems \& Control Letters}, vol.~54, no.~4, pp. 325--329, 2005.

\bibitem{Coulson2019}
J.~Coulson, J.~Lygeros, and F.~Dörfler, ``Data-enabled predictive control: In the shallows of the deepc,'' in \emph{2019 18th European Control Conference (ECC)}, 2019, pp. 307--312.

\bibitem{Pannocchia2001}
G.~Pannocchia and J.~B. Rawlings, ``The velocity algorithm lqr: a survey,'' TWMCC, Department of Chemical Engineering, University of Wisconsin-Madison, Tech. Rep. Technical Report 2001–01, 2001.

\bibitem{Wang2004}
L.~Wang, ``A tutorial on model predictive control: Using a linear velocity-form model,'' \emph{Developments in Chemical Engineering and Mineral Processing}, vol.~12, no. 5-6, pp. 573--614, 2004.

\bibitem{Lazar2022}
M.~Lazar and P.~C.~N. Verheijen, ``Offset–free data–driven predictive control,'' in \emph{2022 IEEE 61st Conference on Decision and Control (CDC)}, 2022, pp. 1099--1104.

\bibitem{Waarde2020}
H.~J. van Waarde, C.~De~Persis, M.~K. Camlibel, and P.~Tesi, ``Willems’ fundamental lemma for state-space systems and its extension to multiple datasets,'' \emph{IEEE Control Systems Letters}, vol.~4, no.~3, pp. 602--607, 2020.

\bibitem{Zhang2023}
K.~Zhang, Y.~Zheng, C.~Shang, and Z.~Li, ``Dimension reduction for efficient data-enabled predictive control,'' \emph{IEEE Control Systems Letters}, vol.~7, pp. 3277--3282, 2023.

\bibitem{Holland2014}
D.~P. Holland, E.~J. Park, P.~Polygerinos, G.~J. Bennett, and C.~J. Walsh, ``The soft robotics toolkit: Shared resources for research and design,'' \emph{Soft Robotics}, vol.~1, no.~3, pp. 224--230, 2014.

\end{thebibliography}

\begin{IEEEbiography}[{\includegraphics[width=1in,height=1.25in,clip,keepaspectratio]{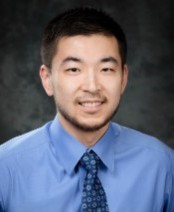}}]{Huanqing Wang}
(Student Member, IEEE) received the B.S. degree in mechanical engineering from the University of Nebraska–Lincoln in 2016 and the M.S. degree in mechanical engineering from Michigan Technological University in 2018. He is currently working toward the Ph.D. degree in mechanical engineering at Michigan State University. His research interests include data-driven control and soft robotics.
\end{IEEEbiography}

\vspace{-20pt}

\begin{IEEEbiography}
[{\includegraphics[width=1in,height=1.25in,clip,keepaspectratio]{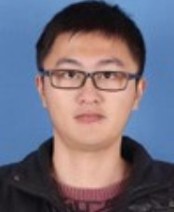}}]{Kaixiang Zhang}
 received the B.E. degree in automation from Xiamen University, Xiamen, China, in 2014, and the Ph.D. degree in control science and engineering from Zhejiang University, Hangzhou, China, in 2019. He was a Research Associate with the Department of Mechanical Engineering, Michigan State University, where he is currently a fixed-term Assistant Professor. His research interests include visual servoing, robotics, and control theory and application.
\end{IEEEbiography}

\vspace{-20pt}

\begin{IEEEbiography}
[{\includegraphics[width=1in,height=1.25in,clip,keepaspectratio]{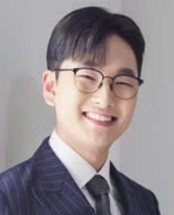}}]{Kyungjoon Lee}
 received the B.S. degree in engineering from the Cooper Union for the Advancement of Science and Art in 2020 and the M.S. degree in mechanical engineering from the University of California, Riverside, in 2025. He is currently working toward the Ph.D. degree in mechanical engineering at the University of California, Riverside. His research interests include soft robotics and wearable assistive technology.
\end{IEEEbiography}

\vspace{-20pt}

\begin{IEEEbiography}
[{\includegraphics[width=1in,height=1.25in,clip,keepaspectratio]{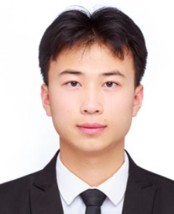}}]{Yu Mei}
(Student Member, IEEE) received the B.S. degree in robotics engineering from the School of Automation, Southeast University (SEU), Nanjing, China, in 2020. He is currently pursuing the Ph.D. degree in electrical and computer engineering with Michigan State University (MSU), East Lansing, MI, USA. His research interests include modeling and control in robotics, soft robotics, rehabilitation robots, and pattern recognition.
\end{IEEEbiography}

\vspace{-20pt}

\begin{IEEEbiography}
[{\includegraphics[width=1in,height=1.25in,clip,keepaspectratio]{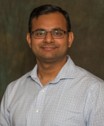}}]{Vaibhav Srivastava}
(Senior Member, IEEE) received the B.Tech. degree in mechanical engineering from the Indian Institute of Technology Bombay, Mumbai, India, in 2007, the M.S. degree in mechanical engineering, the M.A. degree in statistics, and the Ph.D. degree in mechanical engineering from the University of California at Santa Barbara, Santa Barbara, CA, USA, in 2011, 2012, and 2012, respectively. He is currently an Associate Professor of Electrical and Computer Engineering, Michigan State University, East Lansing, MI, USA. His research interests include cyber physical human systems with emphasis on mixed human-robot systems, networked multi-agent systems, aerial robotics, and connected and autonomous vehicles.
\end{IEEEbiography}

\vspace{-20pt}

\begin{IEEEbiography}
[{\includegraphics[width=1in,height=1.25in,clip,keepaspectratio]{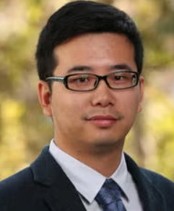}}]{Jun Sheng}
(Member, IEEE) received the B.Sc. degree in mechanical engineering from Shanghai Jiao Tong University, Shanghai, China, in 2011, the M.Sc. degree in electrical engineering from National Taiwan University, Taipei, Taiwan, and the Ph.D. degree in robotics from the Georgia Institute of Technology, Atlanta, GA, USA. He is currently an Assistant Professor of Mechanical Engineering at the University of California, Riverside. His research interests are centered around imaging-guided surgical robotics, medical devices, and smart materials.
\end{IEEEbiography}

\vspace{-20pt}

\begin{IEEEbiography}
[{\includegraphics[width=1in,height=1.25in,clip,keepaspectratio]{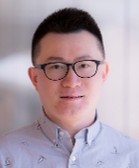}}]{Ziyou Song}
(Senior Member, IEEE) received the bachelor’s and Ph.D. degrees (Hons.) in automotive engineering from Tsinghua University, China, in 2011 and 2016, respectively. He is currently an Assistant Professor at the Department of Electrical Engineering and Computer Science, University of Michigan, Ann Arbor. His research interests include modeling, estimation, optimization, and control of energy storage systems, especially for the electrified transportation sector.
\end{IEEEbiography}

\vspace{-20pt}

\begin{IEEEbiography}
[{\includegraphics[width=1in,height=1.25in,clip,keepaspectratio]{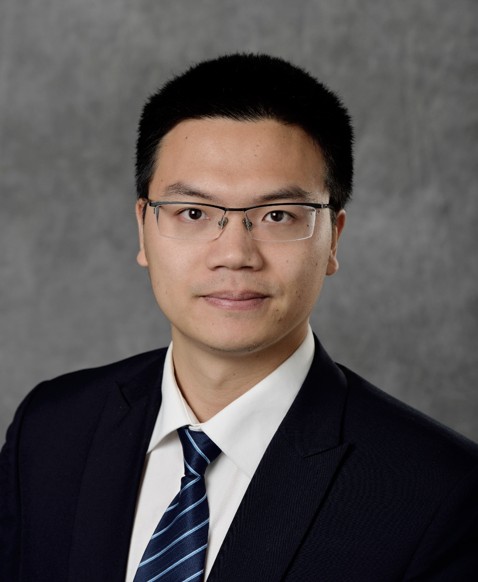}}]{Zhaojian Li}
(Senior Member, IEEE) received the B.Eng. degree from Nanjing University of Aeronautics and Astronautics in 2010 and the M.S. and Ph.D. degrees in aerospace engineering (flight dynamics and control) from the University of Michigan, Ann Arbor, in 2013 and 2015, respectively. He is currently a Red Cedar Distinguished Associate Professor with the Department of Mechanical Engineering, Michigan State University. His research interests include learning-based control, nonlinear and complex systems, and robotics and automated vehicles. He was a recipient of the NSF CAREER Award.
\end{IEEEbiography}

\vspace{-20pt}

\end{document}